\documentclass[letterpaper, 10 pt, conference]{ieeeconf}  

\usepackage{amsmath,amsfonts,amsthm}
\usepackage[noend]{algpseudocode}
\usepackage{graphicx}
\usepackage{makecell}
\usepackage{caption}

\usepackage{subfigure}
\usepackage{xcolor}
\usepackage{multirow}

\IEEEoverridecommandlockouts                              

\title{\LARGE \bf
Interaction-Aware Behavior Planning for Autonomous Vehicles Validated with Real Traffic Data
}

\author{Jinning Li, Liting Sun, Wei Zhan, Masayoshi Tomizuka
\thanks{This work was supported by Momenta.}
\thanks{J. Li, L. Sun, W. Zhan and M. Tomizuka are with the Department of Mechanical Engineering, University of California, Berkeley, CA 94720, USA
        {\tt\small 	\{jinning\_li, litingsun, wzhan, tomizuka\}@berkeley.edu}}%
}

\begin{document}

\maketitle
\thispagestyle{empty}
\pagestyle{empty}

\begin{abstract}

Autonomous vehicles (AVs) need to interact with other traffic participants who can be either cooperative or aggressive, attentive or inattentive. Such different characteristics can lead to quite different interactive behaviors. Hence, to achieve safe and efficient autonomous driving, AVs need to be aware of such uncertainties when they plan their own behaviors. In this paper, we formulate such a behavior planning problem as a partially observable Markov Decision Process (POMDP) where the cooperativeness of other traffic participants is treated as an unobservable state. Under different cooperativeness levels, we learn the human behavior models from real traffic data via the principle of maximum likelihood. Based on that, the POMDP problem is solved by Monte-Carlo Tree Search. We verify the proposed algorithm in both simulations and real traffic data on a lane change scenario, and the results show that the proposed algorithm can successfully finish the lane changes without collisions.

\end{abstract}

\section{INTRODUCTION}

Although rapid progress has been made in autonomous driving in the past decade, there are many challenges yet to be solved. One of such challenges lies in the behavior planning of autonomous vehicles (AVs) in the presence of uncertainties introduced during interactions with other traffic road participants. To safely and efficiently share roads with other road users, AVs need to plan their own behaviors while considering the interactive behaviors of others via behavior prediction  \cite{brown_modeling_2020, zhan_probabilistic_2018}. For instance, an autonomous vehicle should predict and reason about what other vehicles might respond if it accelerates to finish a take-over. A cooperative and attentive driver may yield, while an aggressive or inattentive driver may not. Depending on their different cooperativeness levels, the autonomous vehicles should behave differently to assure safety and efficiency.

Many research efforts have been attracted to the behavior planning problem under interactive uncertainties. 
These approaches can be categorized into three groups. The first group is direct policy learning via reinforcement learning (RL) \cite{pAC, Wang_2019}. For example, \cite{pAC} utilized passive actor-critic (pAC) in RL to generate interactive policies, while \cite{Wang_2019} used Deep Deterministic Policy Gradient for policy learning. The second group is to formulate the interaction-aware behavior planning as a partially observable Markov Decision Process (POMDP). Namely, the uncertainties will first be estimated via observations for better decision-making. Many different hidden variables have been proposed to represent different uncertainties, including sensor noises \cite{Sensorerror}, occlusions \cite{sun_behavior_2019, PaS_Katie, Occlusion_Mykel}, human intentions \cite{hubmann_automated_2018,zhan2016non,sefati_towards_2017,Bandyopadhyay}, and cooperativeness of humans \cite{PGM}, to name a few. To reduce the computation load of POMDP, the authors of \cite{MPDM} proposed a multi-policy Decision Making (MPDM) method to simplify the POMDP problem. 

However, the first two groups do not explicitly consider the mutual influence between the behaviors of AVs and other road users. Namely, in terms of the POMDP formulation, the estimation of the hidden variables (uncertainties) is assumed to depend only on the historical states of interactive agents, i.e., not influenced by the potential behaviors of AVs. Such an assumption, however, fails to capture the fact that the intentions or the cooperativeness of other agents can be influenced by the AVs behaviors since other agents are (approximately) rational agents instead of non-intelligent obstacles. Therefore, the third group formulates the interaction-aware behavior planning problem by explicitly leverage such influence. For example, \cite{Sadigh} formulated the problem as a nested optimization problem assuming that the other agents always respond to the AVs with their best responses. \cite{Gabriel} proposed cooperative behavior planning approaches. In \cite{Peng_2019}, Peng et al. proposed to use Bayesian persuasion model to model the "negotiation" process of interactions. In \cite{Hubmann}, the authors represented the human behaviors via heuristic low-level driver models (intelligent driver model) that is influenced by the potential actions from the AVs. 

All the approaches mentioned above have been verified for effectiveness in one or multiple driving scenarios via simulations. However, most of them have not been verified with real traffic data. In this paper, we aim to propose an interaction-aware behavior planning algorithm with validation on real traffic data. Focusing on a lane-changing scenario, we also formulate the problem as addressed in the third category. More specifically, similar to \cite{Hubmann}, we represent the continuous-time trajectories of the other agents via low-level driver models, and estimate the cooperativeness of humans considering the influence from future actions of the AVs. To generate more human-like interactive behaviors, we learn the optimal distribution of the low-level driver models from data to capture the diversity of different humans. Based on the learned models and the estimated cooperativeness, we solve the behavior planning problem via Monta-Carlo Tree Search (MCTS). 
Finally, the proposed approach is verified in both simulation and real traffic data. The results show that the proposed planner can generate actions for the AVs to successfully execute the lane-changing task. Statistical results on the real traffic data also show that the proposed planner with the learned model can generate more human-like behaviors compared to those with heuristic driver models.

\section{PROBLEM STATEMENT}
Throughout the paper, we focus on the lane-changing scenarios. Three vehicles will be considered: one AV (the ego agent $N_0$) and two surrounding agents on the target lane: one following vehicle $N_1$ and one leading vehicle $N_2$ driving. We discretize the behavior/action space of the ego agent. To model the trajectories of the surrounding agents, we utilize hierarchical representations, similar to \cite{Hubmann}. Namely, we represent their continuous trajectories by a high-level discrete model decision and a collection of stochastic low-level driver models. Details regarding the formulation are given below.

\subsection{Partially Observable Markov Decision Process}
A Partially Observable Markov Decision Process (POMDP) can be defined as a tuple $ <\mathcal{X,A,T,O,Z,R}, b_0, \gamma >$, where $x \in \mathcal{X}$ is the state of the agents, $o \in \mathcal{O}$ is the observation perceived by the agents, and $a\in \mathcal{A}$ is the action taken by the agents. The transition model $\mathcal{T}(x,a,x')$ is the probability of the agents ending in the state $x'$ when in the state $x$ and taking the action $a$. The observation model $\mathcal{Z}(x', a, o)$ is the probability of the agents obtaining the observation $o$ after it executes the action $a$ and ends in the state $x'$. $\mathcal{R}(x,a)$ is the reward of taking the action $a$ in the state $x$.$b_0$ is the initial belief state of the agents, and $\gamma$ is the discount factor. The discount factor makes it possible to favor immediate rewards over rewards in the future. 

Unlike the Markov Decision Process (MDP) where we want to find a mapping $\pi: x \mapsto a$, the policy in a POMDP is a mapping $\pi: b \mapsto a$ where it gives an action $a$ for a given belief state $b$. The overall objective is to maximize the expected cumulative discounted reward and find the corresponding optimal policy 
\begin{equation}
    \pi ^{*} = \mathop{\text{arg max}}_{\textbf{$\pi$}} E \left[ \sum_{t=0}^{\infty} \gamma ^t R(x_t, \pi (b_t))\right]
    \label{eq:POMDP}
\end{equation}
\subsubsection{State Space}
We use Fren\'et Frame to express the position, the velocity, and the acceleration of the three vehicles. For $k=\{0,1,2\}$, the position and the velocity of the vehicle $N_k$ are denoted by $s_k$ and $v_k$, respectively. $d_k$ denotes the distance deviation from $N_k$ to the center line of its lane. $l_k$ is the lane number of the vehicle $N_k$. The state of the ego car is denoted by $x_0 = (s_0, d_0, v_0, l_0)^\top$.
The state of the other car is defined as $x_k = (s_k, v_k, l_k, m_k)^\top, k\in \{1,2\}$, where $m_k$ is a hidden state indicating the conservatism of that car. The hidden state $m_k$ cannot be measured directly, but can be estimated based on observations. The uncertainty in the proposed POMDP comes in here, as we add the hidden state $m_k$ into the state space. If $m_k$ is 1, then that car is willing to yield to the ego car, whereas if $m_k$ is 0, then it will not drive cooperatively, not yielding to the ego car. This hidden state $m_k$ induces a belief state space in POMDP. It affects the structure of the driver model in our algorithm, which predicts the human cars' behavior. Our algorithm then plans the next action according to its prediction. 

\subsubsection{Action Space}
Similar to \cite{Hubmann}, the action space of our framework also consists of two dimensions: the longitudinal acceleration, and the lateral velocity of the ego vehicle. We simplify the actions to discrete actions, aiming to ease the curse of dimension in POMDP. The planned action would be sent to lower level planners to be smoothed so as to ensure the comfort. For the longitudinal acceleration, $A_{long} = \{-1.5m/s^2, 0m/s^2, 1m/s^2\}$. For the lateral velocity, $A_{lat} = \{-0.5m/s, 0m/s, 0.5m/s\}$. Here, we assume the ego car can achieve the lateral velocity instantly (relative to one time step), thus it makes sense to use velocity as actions. 

\subsubsection{Transition model}
The transition model is defined by a set of discrete difference equations as follows:
\begin{align}
    &s_k ' = s_k + v\Delta t + \dfrac{1}{2}a_k\Delta t^2 &k \in \{0, 1,2\} \\
    &v_k ' = v_k + \Delta ta_k &k \in \{0, 1,2\}\\
    &l_k ' = l_k \pm 1 &k \in \{0\}\\
    &l_k ' = l_k &k \in \{1,2\} \\
    &d_k ' = d_kv_{k,lat} + (l_k '-l_k)w_{lane} &k \in \{0\}\\
    &m_k ' = m_k &k \in \{1,2\}
\end{align}
where $w_{lane}$ is the width of the current lane of the cars, and $\Delta t$ is the sampling time. The left hand sides of Eq. 4-9 are values at the next time step. In our problem setting, we assume that the other two human cars would not change their lanes. We want the ego car to change to the lane of the other two cars. Therefore, the action of the ego car $a_0$ is obtained by solving the POMDP, and the actions of the other two human cars are predicted by driver models when solving the POMDP. 

\subsubsection{Observation Space}
 We assume that the ego car has access to all data except for the hidden state $m_k$ from the sensors. Therefore, the observation of the ego car is defined as $o_0 = (s'_0, d'_0, v'_0, l'_0)^\top$.
The observation of the other cars is defined as $o_k = (s'_k, v'_k, l'_k)^\top, k \in \{1,2\}$.
Here, the hidden state $m_k$ is not presented in the observation space. It needs to be estimated by a logistic regression classifier during the planning process. The classifier is presented in the following section. 

\subsubsection{Reward Function}
We adopt the reward function of our framework from the definition in \cite{Hubmann}. The reward is a sum of several different rewards, shown as Eq. \ref{eq:reward}. The parameters were tuned and worked well in practice. 
\begin{gather}
    \begin{aligned}
        r(x,a,x')=&r_{vel}+r_{act}+r_{end\_lane}\\
        &+r_{wrong\_lane}+r_{center}+r_{collision}
    \label{eq:reward}
    \end{aligned}
\end{gather}
$r_{vel}$, which is defined in Eq. \ref{eq:rvel1} and Eq. \ref{eq:rvel2}, denotes the reward for the deviation from the reference velocity. We choose a reference velocity based on traffic rules and vehicle dynamics, and give it to the ego car to follow. 
\begin{align}
    &r_{vel} = -100(v_{ref}-v_0)^2, &\text{if}\:\:\: v_0>v_{ref}
    \label{eq:rvel1}
    \\
    &r_{vel} = -100(v_{ref}-v0), &\text{if}\:\:\: v_0<v_{ref}
    \label{eq:rvel2}
\end{align}
$r_{wrong\_lane}$ is the wrong lane reward. If the ego car is not driving on the target lane, then $r_{wrong\_lane} = -10000$. The end of lane reward $r_{end\_lane}$ encourages the ego car to change its lane before it reaches the end of the road. $r_{end\_lane}$ is negative and it decreases from 0 to -1000 linearly over the last 50m of the road. 

$r_{center}$ is the reward for the deviation from the center line of the target lane. It encourages the ego car to merge into the target lane, instead of driving on its original lane. Thus, it is defined to be $r_{center} = -200d_0^2$.

$r_{act}$ encourages the ego car to take an action that can result in better comfort. It penalizes big value of longitudinal acceleration and latitudinal velocity with $r_{act} = -100(a_{0,long}^2+2|v_{0,lat}|)$.

A collision in autonomous vehicles planning is intolerant, so the reward for collision should be extremely negative, with $r_{collision} = -1000000$.

\subsection{Yielding Classifier}
When solving the POMDP, we need to predict the future actions of the other two human cars. We use popular driver models to make this prediction. Predicting actions of some car often involves choosing a suitable driver model for it, so we need a mechanism to make this choice. We use a yielding classifier to estimate how likely the human car is going to yield to the ego car. Then the suitable driver model for the human car is selected based on the probability of yielding. The yielding classifier is a logistic regression classifier trained by real-world data set. The probability of the human car yielding to the ego car is expressed as Eq. \ref{eq:logistic} and Eq. \ref{eq:classifier}.
\begin{align}
    &h_\theta (f_k) = \dfrac{1}{1+e^{-\theta^Tf_k}}
    \label{eq:logistic}
    \\
    &P_{k,yield}=1, \:\:\text{if}\:\:\: h_\theta (f_k)>\beta_{yield}
    \label{eq:classifier}
\end{align}
where $f_k = [1,\phi_k,d_0,v_0,s_k-s_0,v_k,v_{k,front}]$ is the feature describing the current state, and $\beta_{yield} = 0.85$ is a threshold. Using this yielding classifier, the ego car is able to take its interaction with the human car into account when generating the action for the next time step.

\section{INTERACTION-AWARE BEHAVIOR PLANNING}
The ego car needs to change to the next lane where two human cars are driving. When solving the optimal policy to execute the lane changing, the ego car must consider its interaction with the human cars. In the merging scenario, the back car on the target lane either yields to or ignores the ego car. It is reasonable to assume the driving behavior of the back car follows one driver model when yielding and another model when not yielding. Aiming to find better driving models to predict the driving behavior of human cars, we use real-world data to learn different sets of parameters for two widely used driver models. We then choose a model that fits the real-world data better as the one used in the prediction process, and use another to build the test environment for our algorithm. 

\subsection{Modeling Human-Driven Vehicles}
Driver models estimate the behavior of other cars on the roadway based on the current state of the car and the car immediately in front of it. The driver models predict the acceleration of the back human car in the target lane based on the state of each of the vehicles. We denote the back car's acceleration, $\hat{a}$, as
\begin{equation}
    \hat{a}=f(x;\theta),
    \label{eq:ahat}
\end{equation}
where the function $f(x;\theta)$ denotes the driver model function of the states $x$ and the model parameters $\theta$. Researchers have done plenty of prior works on human vehicles model and prediction with complex algorithm design \cite{Jiachen_ITSC18-2,hu_multi-modal_2019,ma_wasserstein_2019}. However, since this work is focused on the design of the planning module, we adopt Intelligent Driver Model and Velocity Driver Model for simplicity and efficiency.

\subsubsection{Intelligent Driver Model}
The Intelligent Driver Model (IDM) \cite{Treiber_2000} is given by Eqns. \ref{eq:IDM1} and \ref{eq:IDM2}. The model describes the acceleration $a$ of the back car, as a function of the car's velocity $v_{back}$, the reference velocity $v_0$, the difference between the car velocity and the velocity of the car in front $\Delta v=v_{back}-v_{front}$, and the following distance $\varphi=s_{front}+N_{length,front}-s_{back}$. Here, $s_{front}$ is the position of the front car, $s_{back}$ denotes the position of the back car, and $N_{length, front}$ denotes the length of the front car.
\begin{align}
       &a=a_{max}\left[1-\left(\frac{v_{back}}{v_{0}}\right)^{\delta}-\left(\frac{\phi\left(v_{back}, \Delta v\right)}{\varphi}\right)^{2}\right]
    \label{eq:IDM1}\\
    &\phi\left(v_{back}, \Delta v\right)=d_{0}+v_{back} T+\frac{v_{back} \Delta v}{2 \sqrt{a b}}
    \label{eq:IDM2}
\end{align}
 The physical interpretation of the parameters are the minimum following time, $T$, the minimum following distance, $d_0$, the maximum acceleration, $a$, and the braking deceleration, $b$. 

The IDM is represented using the state vector $x$ and parameter vector $\theta$ per Eq. \ref{eq:ahat}, where $x$ and $\theta$ are defined by
\begin{align}
    &x = \begin{bmatrix}v_{back} & \Delta v & \varphi \end{bmatrix}^T\\
    &\theta = \begin{bmatrix}T& a_{max} &v_0 &\delta &d_0& b \end{bmatrix}^T.
\end{align}

\subsubsection{Velocity Difference Model} 
The Velocity Difference Model \cite{Jiang} is given in Eq. \ref{eq:VDM1} and \ref{eq:VDM2}. 
\begin{equation}
    a = \kappa[V(s)-v_{back}+\lambda \Delta v]
    \label{eq:VDM1}
\end{equation}
where $\kappa$ is a sensitivity constant and $V$ is the optimal velocity that the drivers prefer. Researchers adopted an optimal velocity function as 
\begin{equation}
    V(s) = V_1 + V_2 \text{tanh} [C_1 \varphi -C_2]
    \label{eq:VDM2}
\end{equation}
where $\varphi = s_{front}+N_{length,front}-s_{back}$, $\Delta v=v_{back}-v_{front}$. Similar to the setting aforementioned, we use the state vector $x$ and parameter vector $\theta$ as 
\begin{align}
    &x = \begin{bmatrix}v_{back} & \Delta v & \varphi \end{bmatrix}^T\\
    &\theta = \begin{bmatrix}V_1 & V_2 & C_1 & C_2 & \lambda & \kappa \end{bmatrix}^T.
\end{align}

\subsection{Parameters Calibration on Real Traffic Data}
We assume the actual acceleration of the back human car on the target lane is given by the predicted acceleration and additive Gaussian noise (Eq. \ref{eq:noise}).
\begin{equation}
   a = f(x;\theta)+\epsilon\textrm{\qquad}\epsilon\sim\mathcal{N}\left(0, \sigma_{\epsilon}^{2}\right)
\label{eq:noise}
\end{equation}

Equivalently to Eq. \ref{eq:noise}, the behavior is described by Eq. \ref{eq:cond} as a Gaussian probability density function of acceleration $a$ conditional on the states $x$ and parameters $\theta$.

\begin{equation}
    p\left(a | x ; \theta,\sigma_\epsilon\right)=\frac{1}{\sqrt{2 \pi} \sigma_{\epsilon}} e^{-\frac{\left(a-\hat{a}(x,\theta)\right)^{2}}{2 \sigma_{\epsilon}^{2}}}
    \label{eq:cond}
\end{equation}

A sequences of recorded $m$ sets of the state and acceleration,$\{(x^{(1)},a^{(1)}),(x^{(2)},a^{(2)}), ...,(x^{(m)},a^{(m)})\}$, are recorded over time for a back and a front human car from the data.
We employ the following Maximum a Posteriori estimation method to estimate the model parameters, $\theta$, using the recorded values of $x$ and $a$. The method is set up to minimize a cost function $J(\theta;x,a)$ to find the optimal parameters $\theta$ (Eq. \ref{eq:min}).
\begin{equation}
    \theta^* = \underset{\theta}{\textrm{argmin }} J(\theta;x,a)
    \label{eq:min}
\end{equation}

\subsubsection{Maximum Likelihood Estimation}
The maximum likelihood method attempts to maximize the log-likelihood of the parameters $\theta$ given the measured states $x$. For a conditional Gaussian, we find the log-likelihood by taking the log of the likelihood $l(\theta;x,a)=p(a|x;\theta)$ in Eq. \ref{eq:cond}. The log-likelihood is given by Eq. \ref{eq:logli}, where the $C$ represents a constant term that is independent of $\theta$ and $x$ and therefore does not affect the optimization problem. 
\begin{equation}
    \log l(\theta;x,a)= C-\frac{1}{2\sigma_{\epsilon} ^2}\sum_{i=1}^{m}\left(a^{(i)}-\hat{a}(x^{(i)},\theta)\right)^2
    \label{eq:logli}
\end{equation}

\subsection{Calibration Implementation}
\subsubsection{Data Collection}
We use a data set collected by drones from a highway ramp merging scenario \cite{zhan2019constructing}. It contains the position, the velocity and the length of each car in the scene. We process the data set to pick up the data of similar scenarios (an ego car merging to the neighbor lane). 

The data is divided into two sets, successful trials, where the merging car is successfully able to merge between the front and back car, and unsuccessful trials, merging car must fall behind the back car in order to merge into the lane. In the unsuccessful case, the back car is directly behind the front car, so its acceleration is predicted by the IDM/VDM using front car as the front car's state. Conversely, in the successful case, the merging car drives between the front and back car, so the acceleration of the back car is no longer calculated from the front car. Instead, the IDM/VDM use the state of the merging car to predict the acceleration of the back car. 

The sets contain $p= 75$ trials for the unsuccessful case, and $p=115$ trials for the successful case. Each trial contains $m$ time steps of data, which varies depending on the trial (around 5 seconds with 100 recordings per second).

The IDM/VDM model parameters are calibrated separately for each trial, since each trial represents a different car. Here the goal is to find a distribution of the model parameters fitting the set of parameters $\{\theta_1, \theta_2, \theta_3, \dots, \theta_p\}$ which we obtained from maximum likelihood estimation, so that the variation in driver behavior can be captured. For each set (successful/unsuccessful), the parameter estimation is found using Maximum Likelihood Estimation method. Outliers are excluded on the basis of the returned value of the cost function, $J(\theta^*;x,a)$; i.e. if $J(\theta^*;x,a)$ of the trial is above a limit $\alpha$ it is excluded from analysis.  The limit $\alpha$ is set according to $\alpha = Q_3+1.5IQR$, where $Q_3$ represents the third quartile of the cost function evaluation for the trials and $IQR$ represents the respective interquartile range.

\begin{table}[!tbp] 
    \caption{Maximum likelihood parameter estimation results for trials with successful merge.}
    \centering
    \vspace{5pt}
    \begin{tabular}{c|c|c|c|c|c|c}
    \hline
    Model & \multicolumn{6}{c}{IDM}\\
    \hline
    $\theta$ & $T$ & $a_{max}$ & $v_0$ & $\delta$ & $d_0$ & $b$\\
    \hline
    Means & 0.522 & 0.837 & 13.257 & 5.696 & 4.543 & 0.805\\
    Variance & 0.710 & 0.691 & 13.484 & 7.990 & 3.776 & 1.476\\
    \hline
    \hline
    Model & \multicolumn{6}{c}{VDM}\\
    \hline
    $\theta$ & $V_1$ & $V_2$ & $C_1$ & $C_2$ & $\lambda$ & $\kappa$ \\
    \hline
    Means & 4.760 & 5.158 & 1.748 & 3.386 & 1.455 & 0.476 \\
    Variance & 3.293 & 3.390 & 1.945 & 3.389 & 2.136 & 0.950\\
    \hline
    \end{tabular}
    \label{tab:MLEresult_s}
\end{table}

\begin{table}[!tbp] 
    \caption{Maximum likelihood parameter estimation results for trials with unsuccessful merge.}
    \centering
    \vspace{5pt}
    \begin{tabular}{c|c|c|c|c|c|c}
    \hline
    Model & \multicolumn{6}{c}{IDM}\\
    \hline
    $\theta$ & $T$ & $a_{max}$ & $v_0$ & $\delta$ & $d_0$ & $b$\\
    \hline
    Means & 0.958 & 1.421 & 16.885 & 3.426 & 1.281 & 61.907\\
    Variance & 1.995 & 0.769 & 15.430 & 2.191 & 2.484 & 483.36\\
    \hline
    \hline
    Model & \multicolumn{6}{c}{VDM}\\
    \hline
    $\theta$ & $V_1$ & $V_2$ & $C_1$ & $C_2$ & $\lambda$ & $\kappa$ \\
    \hline
    Means & 3.747 & 6.133 & 1.641 & 7.118 & 0.530 & 0.332 \\
    Variance & 3.507 & 3.433 & 2.289 & 3.528 & 0.514 & 0.388\\
    \hline
    \end{tabular}
    \label{tab:MLEresult_ns}
\end{table}

\subsubsection{Parameter Estimation Results}
We assume the parameters are independently distributed. For each trial in the data set, we learn a set of parameters using Maximum Likelihood Estimation (MLE) method. Then we collect all sets of parameters and fit them into a Gaussian distribution. 
We learn parameters for both successful merging cases and unsuccessful merging cases, as the data set consists of these two kinds of scenarios. The resulting parameters distributions (mean and covariance) of $\theta$ are shown in table \ref{tab:MLEresult_s} and table \ref{tab:MLEresult_ns}. We compare the accuracy of the predicted values using two driver models: IDM, and VDM. We performed the estimation using MLE for each of the driver models. Table \ref{tab:MSE} gives the mean square error (MSE) between the measured acceleration and predicted acceleration statistics for all of the trials for the MLE method. The average and maximum MSE over all the trials is the least for the VDM, following by the IDM. Notice that the standard deviation of VDM parameters is less than that of IDM parameters in both successful and unsuccessful merging scenarios. Thus, we conclude that within the scenarios described in our problem setting, the tuned VDM model achieves a better performance than the tuned IDM model. Hence, we choose to use the tuned VDM model as the prediction model in the POMDP framework. When the algorithm is making a prediction of the motion of other cars, the VDM parameters would switch between two sets (either $\theta_s$ or $\theta_{ns}$) according to the output of the yielding classifier. In addition, we also change the front car in the driver model using the output of the yielding classifier. If the probability of yielding is 1, then the front car is set to be the ego car, otherwise it is set to be the lead human car on the target lane. 

\begin{table}[!tbp] 
    \caption{Average and maximum (over all trials) mean-squared error for each parameter estimation method.}
    \centering
    \vspace{5pt}
    \begin{tabular}{c|c|c}
    \hline
      Model   & Average MSE & Max MSE \\
      \hline
        VDM &0.046&0.139\\
        IDM &0.072&0.451\\
         \hline
    \end{tabular}
    \label{tab:MSE}
\end{table}

\section{IMPLEMENTATION}
The overall goal of our algorithm is to find an optimal policy for the agent to execute at the next time step. In our problem setting, the aim is to generate a best longitudinal acceleration and lateral velocity for the ego agent, so that a few seconds later it can merge into the next lane safe and sound. One of the main drawbacks of POMDP is the curse of dimension. The computational burden will increase exponentially as the rollout horizon increases. To ease this situation, we apply Monte-Carlo Tree Search (MCTS) to search a pair of best action $a = \mathcal{A}_{long}\times \mathcal{A}_{lat}$ for the next time step. We do MCTS over a 10 seconds horizon to solve the POMDP. 

\subsection{Monte-Carlo Tree Search}
Monte-Carlo Tree Search (MCTS) is an efficient search technique, which can balance the trade-off between exploration and exploitation. When we use a tree search, it is very likely that the current perceived best action is not the global optimal action. MCTS is good at dealing with this problem. It is able to search the unexplored nodes in the tree while also be favor of the area where better actions exist. Therefore, MCTS can result in relatively good action without a traverse of the tree. Actually, as the number of iterations increase, the result of MCTS converges to the real optimal action in terms of probability \cite{Kurniawati}. This property can reduce the computational burden when solving our POMDP framework, and generate a best action for the autonomous vehicle to execute at the next time step within a relatively small time period. 

\begin{figure*}[t]
    \centering
        \begin{subfigure}
        \centering
    \includegraphics[width=0.31\textwidth]{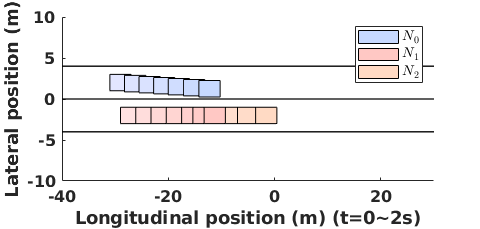}
        \end{subfigure}
    \begin{subfigure}
        \centering
    \includegraphics[width=0.31\textwidth]{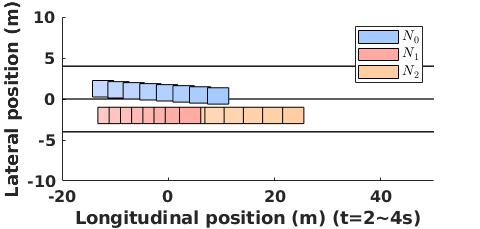}
        \end{subfigure}
    \begin{subfigure}
        \centering
    \includegraphics[width=0.31\textwidth]{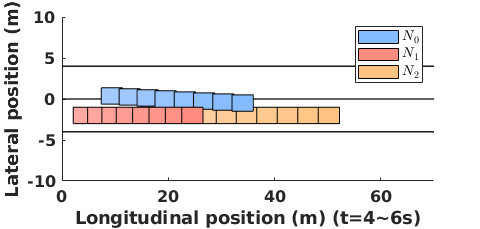}
        \end{subfigure}
    \begin{subfigure}
        \centering
    \includegraphics[width=0.31\textwidth]{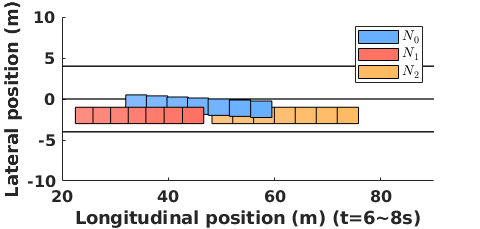}
        \end{subfigure}
    \begin{subfigure}
        \centering
    \includegraphics[width=0.31\textwidth]{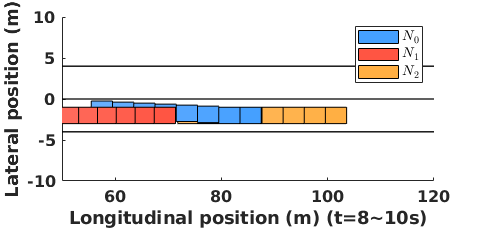}
        \end{subfigure}
    \begin{subfigure}
        \centering
    \includegraphics[width=0.31\textwidth]{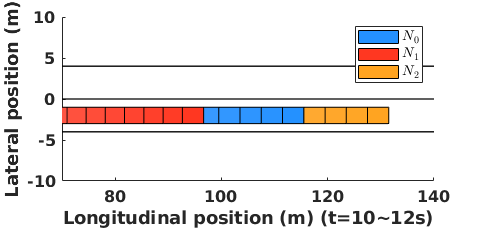}
        \end{subfigure}
        \caption{The trajectory of the ego agent when the back human car is yielding to it. The continuous trajectory is sliced into 6 pieces for clarity. Each trajectory piece lasts 2 second and the whole trajectory lasts 12 seconds.}
        \label{fig:yield}
\end{figure*}

\begin{figure*}[t]
    \centering
        \begin{subfigure}
        \centering
    \includegraphics[width=0.31\textwidth]{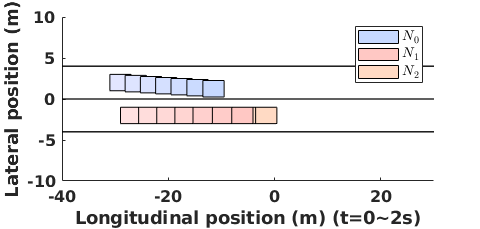}
        \end{subfigure}
    \begin{subfigure}
        \centering
    \includegraphics[width=0.31\textwidth]{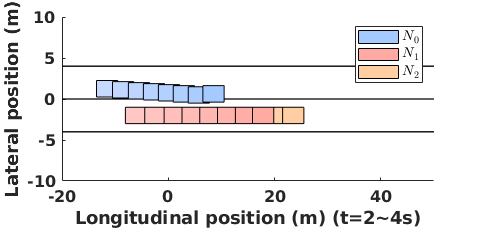}
        \end{subfigure}
    \begin{subfigure}
        \centering
    \includegraphics[width=0.31\textwidth]{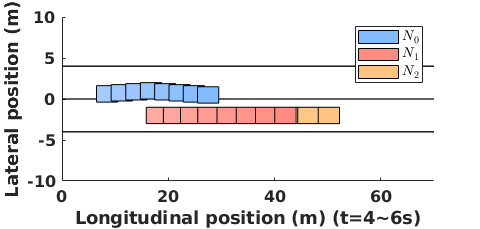}
        \end{subfigure}
    \begin{subfigure}
        \centering
    \includegraphics[width=0.31\textwidth]{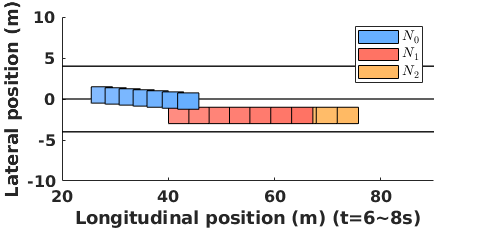}
        \end{subfigure}
    \begin{subfigure}
        \centering
    \includegraphics[width=0.31\textwidth]{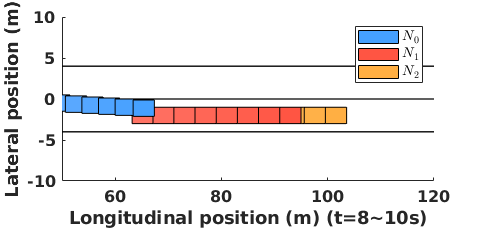}
        \end{subfigure}
    \begin{subfigure}
        \centering
    \includegraphics[width=0.31\textwidth]{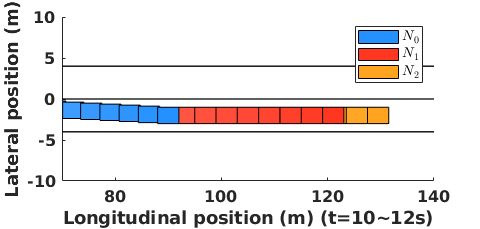}
        \end{subfigure}
        \caption{The trajectory of the ego agent when the back human car is yielding to it. The continuous trajectory is sliced into 6 pieces for clarity. Each trajectory piece lasts 2 second and the whole trajectory lasts 12 seconds.}
        \label{fig:not_yield}
\end{figure*}

\begin{figure}[t]
    \centering
    \begin{subfigure}
        \centering
        \includegraphics[width=0.45\textwidth]{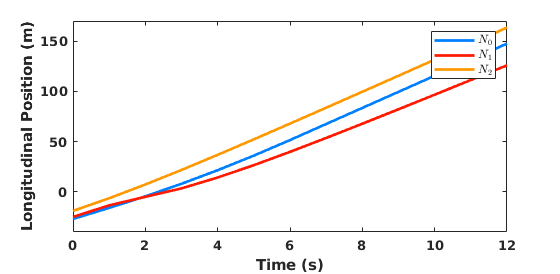}
    \end{subfigure}
    \begin{subfigure}
        \centering
        \includegraphics[width=0.45\textwidth]{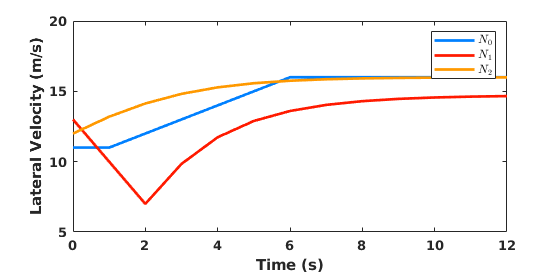}
    \end{subfigure}
    \caption{The longitudinal position and velocity of the cars when the back human car is yielding to the ego agent.}
    \label{fig:vel_yield}
\end{figure}
\begin{figure}[t]
    \centering
    \begin{subfigure}
        \centering
        \includegraphics[width=0.45\textwidth]{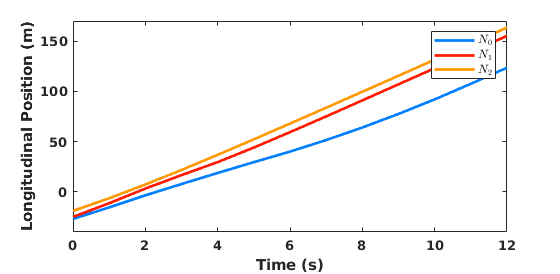}
    \end{subfigure}
    \begin{subfigure}
        \centering
        \includegraphics[width=0.45\textwidth]{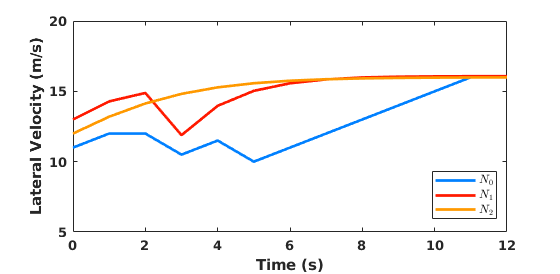}
    \end{subfigure}
    \caption{The longitudinal position and velocity of the cars when the back human car is not yielding to the ego agent.}
    \label{fig:vel_notyield}
\end{figure}

\subsection{Rollout Details}
When we implement MCTS, we need to do rollout over a 10 seconds lookahead. In the process of rollout, we have to predict the actions of other human cars, so that we can get the expected reward over the 10 seconds horizon. As aforementioned, we use Velocity Driver Model (VDM) to generate the motion of other cars in the prediction module when we do the rollout. The yielding classifier first takes in the feature and outputs a probability of cooperative behavior of the other car. Then if the probability of yielding to the ego car is bigger than the threshold, the front car in VDM is set to be the ego car, and the parameters of VDM are set to be those learned from successful merging scenarios. If the probability of yielding to the ego car is less than the threshold, then the front car in VDM is set to be the lead human car on the target lane, and the parameters of VDM are set to be those learned from unsuccessful merging scenarios. This setting makes sense because the behavior patterns of a human car would be different between the cooperative and uncooperative scenarios. In this way, the ego car is aware of the intention of the other human cars, and is able to make decisions in consideration of its interaction with other cars. 


\section{EXPERIMENTS}
We implemented the distribution for the parameters found using MLE into an POMDP described in the Problem Setting Section. We used MATLAB to carry out the experiments on a system with a 2.9GHz Intel Core i7-7500U CPU. 

\subsection{Simulation Results}
In the experiment, we set the horizon to be $12$ seconds. There were totally three cars: one was the ego car in lane 1, intending to merge into lane 0, and the other two cars were the other human cars in lane 0. The ego car used POMDP to obtain its action for the next time step. The time step was $1$ second. When solving MCTS for POMDP, we used IDM/VDM and a yield classifier to predict the motion of the other cars. IDM was a baseline whose parameters were set according to widely accepted expert data from previous literatures. Before starting each trial, we sampled parameters for VDM from the learned Gaussian distribution of successful and unsuccessful merge. The solver had a $10$-second look-ahead to search for the actions with the maximum reward. Then the ego car executed the best action returned from the search for the next time step. 

Those two other human cars were assumed to be able to change their longitudinal acceleration only, staying at the center line of lane 0. We used the VDM with sampled parameters from the learned distribution by MLE to steer the other cars in the test environment to interact with the ego car.  

We compared the performance of using IDM or VDM as the predictor in the rollout based on 500 trials for each. The ego car can successfully merge into the target lane with a rate of $99.4\%$ when using the VDM to predict the other cars' motion. However, the performance of original IDM as a predicted model for the other cars was not as good, with a success rate as $87.4\%$. Besides, the error of the predicted acceleration, velocity and position of the other cars were also small when using VDM, indicating that the VDM with a set of tuned parameters can better describe the motion of the other cars in the real world. Those results are shown in Table \ref{tab:dataresult}.

In the following paragraph, two trials of merging are discussed: one is in a scenario where the rear human car $N_1$ yielded to the ego agent, while in the other scenario $N_1$ did not yield. These two trials began with the same initial conditions and are implemented in a time horizon of $t = 12s$. The ego car was able to recognize the intentions of the human car using the yield classifier and the tuned prediction model, and executed the lane changing task successfully whether or not the human car yielded. 

In Figure \ref{fig:yield}, we show a trial in which the rear human car yielded to the ego car. The position and the velocity of the three cars are shown in Figure \ref{fig:vel_yield}. As the figures show, $N_1$ in lane 0 decelerated when the ego car cut in, indicating that they were interacting with each other. Meanwhile, the other human car $N_2$ was accelerating to the reference velocity $16m/s$ as it was the speed limit. The ego car was able to find the right time point (around $t=2s$) to execute the merging attempt and then can successfully merge into the target lane as shown in Figure \ref{fig:yield}. In Figure \ref{fig:not_yield}, we show a trial in which the rear human car did not yield to the ego car. The position and the velocity of the three cars are shown in Figure \ref{fig:vel_notyield}. As shown in the velocity plot in Figure \ref{fig:vel_notyield}, the ego car $N_0$ accelerated and attempted to merge at around $t = 0~1s$. However, $N_1$ also accelerated, which means that it did not intend to yield to the ego car $N_0$. Thus, the ego car $N_0$ decelerated a little bit, and then merged into the target lane, behind those two human cars $N_1$ and $N_2$. After successfully merging into the target lane, the ego car $N_0$ then accelerated to the reference speed $v_{ref}=16m/s$, and followed those two human cars afterwards. 

\begin{table}[!tbp]
    \caption{The prediction error and the success merges of IDM/VDM.}
    \centering
     \begin{tabular}{c|c|c|c|c}
     \hline
        \multicolumn{5}{c}{Simulation}\\
        \hline
        model  & a error&v error & x error & success \\
        \hline
        IDM & $0.610$ m/s$^2$ & $1.261$ m/s & $0.630$ m & $437/500$\\
        VDM& $0.483$ m/s$^2$ & $1.134$ m/s & $0.535$ m & $497/500$\\
         \hline
         \hline
         \multicolumn{5}{c}{Real Traffic Data}\\
         \hline
         model  & a error&v error& x error & success\\
        \hline
        VDM & $0.606$ m/s$^2$ & $1.783$ m/s & $0.891$ m & $190/200$\\
        \hline
    \end{tabular}
    \label{tab:dataresult}
\end{table}

\subsection{Real-World Dataset Results}
We utilized the INTERACTION dataset \cite{interactiondataset} to validate the proposed algorithms. Similar to the setup in the calibration implementation, we selected the motion data of similar merging scenarios from the dataset, and used the selected data as the validation dataset. Each scenario in the validation set contains the initial position and trajectories of the ego car, as well as trajectories of two other human cars that are driving on the lane adjacent to the initial lane of the ego car. The goal of the ego car was to merge to the lane with two other cars. The scenarios of the real-traffic dataset were very similar to the simulation in the previous section. We selected 200 merging scenarios from the original dataset, and perform lane change behavior planning in each of them, which is similar to the tasks in the simulation. The statistical results are shown in table \ref{tab:dataresult}. We define a "success" in behavior planning as that a collision-free trajectory is generated. Specifically it means that there is no intersection between the ego car’s trajectory and the other cars’ trajectories at each time step. There were 190 collision free merging attempts out of 200 planning trials, with an average error of 0.606 $m/s^2$ in acceleration prediction, an average error of $1.783m/s$ in velocity prediction, and an average error of $0.891m$ in position prediction. All performance indices including a success rate, were lower than those in simulation, which indicates it is harder to predict the motion of a real human car. However, the proposed planner was still able to perform a merging maneuver with a relatively high success rate in real-world scenarios.

\section{CONCLUSION}

In this paper, we presented a POMDP framework to cope with behavior planning in a lane changing scenario, while considering the interaction between the autonomous car and the surrounding human cars. In the planning module, a yielding classifier was trained by a real-world dataset, predicting intentions and behavior of the human cars by switching the driver model. The switching process was based on the current and historical states, which indicates how likely the human car will drive cooperatively. The driver model switched between two sets of parameters for successful and unsuccessful merges, respectively, learned from real-world data. Evaluations on simulation showed that the proposed framework with learned parameters is able to perform safe and efficient behavior planning for autonomous cars in the lane changing scenarios. We also validated our planner with real-world traffic data environment. The proposed planner with refined prediction model outperformed the planner with a fixed-parameter driver model. In the future, we propose to incorporate various persuasion model to the POMDP framework, so as to make the autonomous car further interact with the surrounding cars, which could result in more efficient and safer actions generated by the behavior planner.

\addtolength{\textheight}{-13cm}   


\bibliographystyle{ieeetr}
\bibliography{reference.bib}


\end{document}